\documentclass{article}

    \usepackage[preprint]{neurips_2020}

\usepackage{amsmath,amsfonts,bm}

\def\eqref#1{equation~\ref{#1}}

\def\1{\bm{1}}

\DeclareMathAlphabet{\mathsfit}{\encodingdefault}{\sfdefault}{m}{sl}
\SetMathAlphabet{\mathsfit}{bold}{\encodingdefault}{\sfdefault}{bx}{n}

\usepackage{xcolor}
\usepackage{multirow}
\usepackage[utf8]{inputenc} 
\usepackage[T1]{fontenc}    
\usepackage{hyperref}       
\usepackage{natbib}
\setcitestyle{numbers}
\usepackage{url}            
\usepackage{booktabs}       
\usepackage{amsfonts}       
\usepackage{nicefrac}       
\usepackage{microtype}      
\usepackage{graphicx}
\usepackage{caption}
\usepackage{subcaption}
\newcommand{\MA}[0]{{\mathcal{A}}}
\newcommand{\MB}[0]{{\mathcal{B}}}
\newcommand{\VA}[0]{\mathbf{\tilde{V}}_{\MA}}
\newcommand{\VB}[0]{\mathbf{\tilde{V}}_{\MB}}
\newcommand{\PA}[0]{\mathbf{P}_{\MA}}
\newcommand{\PB}[0]{\mathbf{P}_{\MB}}
\newcommand{\NA}[0]{N_{\MA}}
\newcommand{\NB}[0]{N_{\MB}}
\newcommand{\TAB}[0]{\mathbf{T}_{\MA \MB}}
\title{WeightedPose: Generalizable Cross-Pose Estimation via Weighted SVD}

\author{
  Xuxin Cheng, Heng Yu, Harry Zhang, Wenxing Deng\\
  Course Project for VLR at CMU, Spring 2023
}

\begin{document}

\maketitle
\begin{abstract}
    We introduce a new approach for robotic manipulation tasks in human settings that necessitates understanding the 3D geometric connections between a pair of objects. Conventional end-to-end training approaches, which convert pixel observations directly into robot actions, often fail to effectively understand complex pose relationships and do not easily adapt to new object configurations. To overcome these issues, our method focuses on learning the 3D geometric relationships, particularly how critical parts of one object relate to those of another. We employ Weighted SVD in our standalone model to analyze pose relationships both in articulated parts and in free-floating objects. For instance, our model can comprehend the spatial relationship between an oven door and the oven body, as well as between a lasagna plate and the oven. By concentrating on the 3D geometric connections, our strategy empowers robots to carry out intricate manipulation tasks based on object-centric perspectives \footnote{\url{https://github.com/harryzhangOG/weighted-pose/tree/main}}.
\end{abstract}

\section{Introduction}
Numerous robotic tasks involve relocating an object to a position relative to another object. For instance, a culinary robot might be required to put a lasagna in an oven, set a pot on a stove, place a dish in a microwave, position a mug on a rack, or situate a cup on a shelf. Mastery in positioning objects according to task-specific needs is crucial for robots working in human environments. Moreover, this capability should extend to new items within known categories, such as inserting different trays into an oven or different mugs onto a rack.

Traditionally in robot learning, policies are trained end2end to translate pixel data directly into robot actions. However, these end-to-end policies struggle with understanding intricate pose relationships like those mentioned and often fail to adapt to new object configurations. In response, we suggest a method that focuses on understanding the 3D geometric relationships between pairs of objects. For the aforementioned tasks, this involves discerning the spatial relationships between critical parts of one object in relation to another—for example, analyzing both the connection between an oven door and the main oven body and between a lasagna plate and the oven itself. Our proposed independent model uses Weighted SVD to analyze these pose relationships, whether they involve articulated parts or free-floating objects.

\section{Background}
\textbf{Object Pose Estimation}:
Pose estimation involves detecting and determining the 6DoF (six degrees of freedom) pose of an object, which encompasses its position and orientation relative to a predefined object reference frame. Recent research has suggested using 3D semantic keypoints as a novel object representation form~\cite{lowe1999object,rothganger20063d,xiang2017posecnn,he2020pvn3d,he2021ffb6d,turpin2021gift}. Recent work~\cite{manuelli2019kpam,qin2020keto, vecerik2021s3k,manuelli2021keypoints} Although keypoint-based methods show generalization within object classes, they rely heavily on extensive hand-annotated data or simulated environments to learn keypoint locations. In contrast, our method can learn from merely 10 real-world demonstrations. Another technique involves using dense embeddings, such as Dense Object Nets (DON) and Neural Descriptor Fields (NDF), which facilitate class-wide generalization by predicting and matching dense embeddings from observations to those of demonstration objects. In contrast, our method is able to learn from just 10 real-world demonstrations. Another approach is to use dense embeddings, such as Dense Object Nets (DON)~\cite{florence2018dense} and Neural Descriptor Fields (NDF)~\cite{simeonov2021neural, devgon2020orienting},
which achieve generalization across classes by predicting dense embeddings in the observation and matching them to embeddings of the demonstration objects. However, both DON and NDF presuppose that the target object is manipulated relative to a static reference in a "known canonical configuration" (for example, the fixed pose of a mug rack in NDF). Unlike these methods, our approach understands the geometric relationships between pairs of objects, allowing it to function without assuming a static environment~\cite{simeonov2021neural}. In contrast, our method reasons about the 
geometric relationship between a pair of objects and hence does not need to assume a static environment.

\textbf{Articulated Object Manipulation}: The manipulation of articulated objects and those with non-rigid characteristics continues to be a challenging area of research, largely because of their intricate geometries and kinematics. Earlier research has suggested handling these objects using hand-crafted analytical methods, for example, by immobilizing a sequence of hinged objects through \cite{Cheong2007-iw, shen2024diffclip, avigal20206, avigal2021avplug, yao2023apla}.
 \citet{berenson2011task} proposed a planning framework for manipulation under kinematic constraints. \citet{Katz2008-jo} proposed an approach to acquire manipulation policies in real-world settings by using a relational representation that is grounded and learned through interaction. This method was enhanced by the creation of expansive datasets of articulated objects like the PartNet dataset \citet{Mo2019-az} and Partnet-Mobility by \citet{Xiang2020-oz, sim2019personalization, jin2024multi}, everal studies have suggested using learning techniques that combine large-scale simulation with supervised visual learning. \citet{Mo2021-jm}introduced an approach to develop articulation manipulation policies through extensive simulation and visual affordance learning. \citet{Xu2021-iw} suggested a framework that acquires knowledge of articulation affordances and includes an action scoring module, which is utilized to manipulate objects. Various studies have specifically concentrated on the visual identification and calculation of articulation parameters, developing methods to estimate the pose \cite{zhang2016health, zhang2023flowbot++, zhang2020dex, Yan2020-hm, Wang2019-gy, Hu2017-bn, Li2020-go} and identify articulation parameters \cite{Jain2021-rg, Zeng2020-tk, zhang2021robots} to obtain action trajectories. Moreover, \cite{Narayanan2015-mp, Burget2013-nb, Chitta2010-vn, lim2021planar, lim2022real2sim2real, elmquist2022art} tackle the problem using statistical motion planning. 

\section{Method}
\begin{figure}
    \centering
    \includegraphics[width=\textwidth]{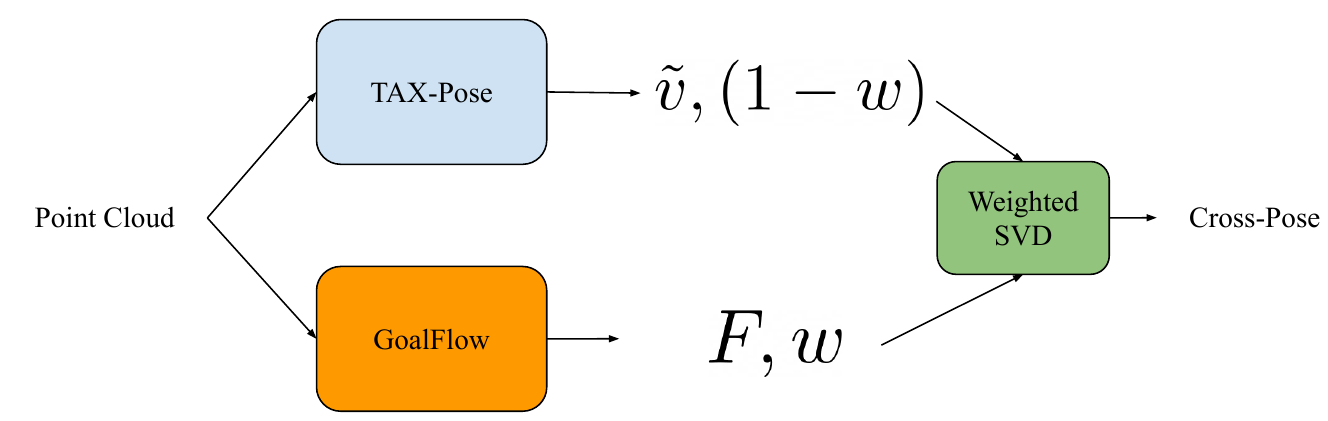}
    \caption{Unified Weighted Pose architecture. The model first takes as input a point cloud, and then learns to predict a weight for the point cloud. This weight is used in the downstream SVD module to combine the GoalFlow and TAX-Pose outputs.}
    \label{fig:weightedpose}
\end{figure}

We first slightly modify the FlowBot 3D training objective. Instead of defining an instantaneous motion vector field as the training data directly, we make the network learn to output a flow field to the completely open state directly. Since this version of FlowBot 3D learns to output a dense representation to the goal state directly, we call this model Goal Flow.

Formally, given a point cloud $\{\mathbf{p}_i\}\;\forall i \in \{1,\dots,N\}$ this Goal Flow model outputs a dense flow field $F\in \mathbb{R}^{N\times 3}$, where each flow vector $\delta_i\in\mathbb{R}^3$ in the $F$ represents a goal flow vector such that point $p_i + \delta_i$ is in the fully open goal state. Ideally, this model should be deployed exclusively for articulated objects in that Goal Flow was shown to achieve suboptimal performances in \cite{pan2022tax, pan2023tax}.

To combine the Goal Flow model with TAX-Pose, we also make the Goal Flow network output an auxiliary weight $w\in \mathbb{R}$, which assigns weight $w$ to Goal Flow and $1-w$ to TAX-Pose. 

For the TAX-Pose component of the Weighted Pose unified architecture, we do not make any significant modifications and ideally the TAX-Pose model should be deployed for free-floating objects exclusively. 

In TAX-Pose, we have:
\begin{equation*}
    \mathcal{J}(\TAB) = \sum_{i=1}^{\NA} \alpha_i^{\MA} || \TAB~\mathbf{p}_{i}^{\MA} - \tilde{\mathbf{v}}_i^{\MA} ||_2^2 + \sum_{i=1}^{\NB} \alpha_i^{\MB} || \TAB^{-1}~\mathbf{p}_{i}^{\MB} - \tilde{\mathbf{v}}_i^{\MB} ||_2^2,
\end{equation*}
where 
\[
    \mathbf{A} = \begin{bmatrix} \PA^{*\top} & \VB^{*\top} \end{bmatrix}, \;\;
    \mathbf{B} = \begin{bmatrix} \VA^{*\top} & \PB^{*\top} \end{bmatrix}^\top, \;\;
  \boldsymbol\Gamma = \text{diag} \left(\begin{bmatrix}  \boldsymbol{\alpha}_{\MA} & \boldsymbol{\alpha}_{\MB} \end{bmatrix}\right)
\]

Now we want to add another term in the SVD step. Specifically, we want the network to learn which one of the two models, TAX-Pose and Goal-Flow, is more important based on point cloud observations. Thus, we want a SVD step that incorporates the TAX-Pose residual, weighted by $(1-w)$ and Goal-Flow weighted by $w$. So the new SVD formulation becomes:
\begin{align*}
    \mathcal{J}(\TAB) &= (1-w)\left[\sum_{i=1}^{\NA} \alpha_i^{\MA} || \TAB~\mathbf{p}_{i}^{\MA} - \tilde{\mathbf{v}}_i^{\MA} ||_2^2 + \sum_{i=1}^{\NB} \alpha_i^{\MB} || \TAB^{-1}~\mathbf{p}_{i}^{\MB} - \tilde{\mathbf{v}}_i^{\MB} ||_2^2\right]\\
    &+w \sum_{i=1}^{\NA}|| \TAB~\mathbf{p}_{i}^{\MA} - (\mathbf{p}_{i}^{\MA} + \delta^\mathcal{A}_i) ||_2^2,
\end{align*}

where $\delta^\mathcal{A}_i$ is the $i-$th point's goal flow. To make this approach viable we optimize $R$ and $t$ in $\TAB$ separately:
\begin{align*}
J &= \sum_i^{N_\mathcal{A}} \left((1-w) \alpha_i \|R \mathbf{p}_i^\mathcal{A} + t - \mathbf{\tilde{v}}_i^\mathcal{A} \|^2 + w \|R \mathbf{p}_i^\mathcal{A} + t - \mathbf{p}_i^\mathcal{A} +\delta_i^\mathcal{A}\|^2 \right) \\
 &+ \sum_i^{N_\mathcal{B}} (1-w) \alpha_i \|R^{-1}\mathbf{p}_i^\mathcal{B} - t - \mathbf{\tilde{v}}_i^\mathcal{B}\|^2
\end{align*}

We first solve for the optimal translation $t^*$:
\begin{align*}
\frac{\partial J}{\partial t} &= 0\\
&= \sum_i^{N_\mathcal{A}}\left(2(1-w)\alpha_i\left(t + R \mathbf{p}_i^\mathcal{A} - \mathbf{\tilde{v}}_i^\mathcal{A}\right) + 2w\left(t+R \mathbf{p}_i^\mathcal{A} - \mathbf{p}_i^\mathcal{A} -\delta_i^\mathcal{A} \right)\right)\\
&+ 2(1-w)\sum_i^{N_\mathcal{B}}\alpha_i\left(R^{-1} \mathbf{p}_i^\mathcal{B} - t - \mathbf{\tilde{v}}_i^\mathcal{B}\right)\\
&= \sum_i^{N_\mathcal{A}}  \left((2-2w)\alpha_i + 2w\right)t + \left((2-2w)\alpha_i + 2w\right) R \mathbf{p}_i^\mathcal{A} - 2w\delta_i^\mathcal{A} - (2-2w)\alpha_i \mathbf{\tilde{v}}_i^\mathcal{A} - 2w\mathbf{p}_i^\mathcal{A}\\
&+ \sum_i^{N_\mathcal{B}} -(2-2w)\alpha_i t + (2-2w)\alpha_i( R^{-1} \mathbf{p}_i^\mathcal{B} - \mathbf{\tilde{v}}_i^\mathcal{B})\\
&= \left[ \sum_i^{N_\mathcal{A}} [(2-2w)\alpha_i + 2w] +   \sum_i^{N_\mathcal{B}} -(2-2w)\alpha_i \right] t \\
&+ \sum_i^{N_\mathcal{A}}\left((2-2w)\alpha_i + 2w\right) R \mathbf{p}_i^\mathcal{A} - 2w\delta_i^\mathcal{A} - (2-2w)\alpha_i \mathbf{\tilde{v}}_i^\mathcal{A} - 2w\mathbf{p}_i^\mathcal{A}\\
&+ \sum_i^{N_\mathcal{B}} (2-2w)\alpha_i( R^{-1} \mathbf{p}_i^\mathcal{B} - \mathbf{\tilde{v}}_i^\mathcal{B})\\
t^* &= -\bigg(\sum_i^{N_\mathcal{A}}\left((2-2w)\alpha_i + 2w\right) R \mathbf{p}_i^\mathcal{A} - 2w\delta_i^\mathcal{A} - (2-2w)\alpha_i \mathbf{\tilde{v}}_i^\mathcal{A} - 2w\mathbf{p}_i^\mathcal{A}\\
&+ \sum_i^{N_\mathcal{B}} (2-2w)\alpha_i( R^{-1} \mathbf{p}_i^\mathcal{B} - \mathbf{\tilde{v}}_i^\mathcal{B})\bigg) \bigg/  \left[ \sum_i^{N_\mathcal{A}} [(2-2w)\alpha_i + 2w] +   \sum_i^{N_\mathcal{B}} -(2-2w)\alpha_i \right]
\end{align*}

Further simplifying, we have:
\begin{align*}
    t^* &= \frac{
    \color{red}(1-w)\sum_i^{N_\mathcal{A}}\alpha_i^\mathcal{A}(\mathbf{\tilde{v}}_i^\mathcal{A} - R \mathbf{p}_i^\mathcal{A})}{\color{red}(1-w)\sum_i^{N_\mathcal{A}}\alpha_i^\mathcal{A} + \color{blue}w\sum_i^{N_\mathcal{A}}1 + \color{purple}(1-w)\sum_i^{N_\mathcal{B}}\alpha_i^\mathcal{B}}\\
    &+ \frac{\color{blue}w\sum_i^{N_\mathcal{A}} (\mathbf{p}_i^\mathcal{A} + \delta_i^\mathcal{A}) - R \mathbf{p}_i^\mathcal{A}}{\color{red}(1-w)\sum_i^{N_\mathcal{A}}\alpha_i^\mathcal{A} + \color{blue}w\sum_i^{N_\mathcal{A}}1 + \color{purple}(1-w)\sum_i^{N_\mathcal{B}}\alpha_i^\mathcal{B}}\\
    &+ \frac{\color{purple}(1-w)\sum_i^{N_\mathcal{B}}\alpha_i^\mathcal{B}(\mathbf{\tilde{v}}_i^\mathcal{B} - R^{-1} \mathbf{p}_i^\mathcal{B})
    }{\color{red}(1-w)\sum_i^{N_\mathcal{A}}\alpha_i^\mathcal{A} + \color{blue}w\sum_i^{N_\mathcal{A}}1 + \color{purple}(1-w)\sum_i^{N_\mathcal{B}}\alpha_i^\mathcal{B}}
\end{align*}
Note here that we colorcode the expression here. Intuitively, the resulting translation is a weighted sum of three translation terms. The \textcolor{red}{red} color represents the action object's translation via TAX-Pose, the \textcolor{blue}{blue} color represents the action object's translation via GoalFlow, and the \textcolor{purple}{purple} color represents the anchor object's translation via TAX-Pose.

We can then construct the matrices as follows:
\begin{equation*}
    \mathbf{A} = \begin{bmatrix} \PA^{\top} & \VB^{\top} & \PA^{ \top}\end{bmatrix}, \;\;
    \mathbf{B} = \begin{bmatrix} \VA^{\top} & \PB^{\top} & \PA^{ \top} + \Delta_\mathcal{A}\end{bmatrix}^\top
\end{equation*}

\begin{equation*}
  \boldsymbol\Gamma = \begin{bmatrix}  (1-w)\cdot\boldsymbol{\alpha}_{\MA}&0&0\\ 0&(1-w)\cdot\boldsymbol{\alpha}_{\MB} &0\\0&0& w\end{bmatrix}
\end{equation*}
where $\Delta_\mathcal{A}$ is the de-meaned goal flow field. Note here that everything here is \textbf{NOT} de-meaned.

We then solve for the SVD:
\[
\boldsymbol{U\Sigma V^\top} = \text{svd}(\boldsymbol{A\Gamma B^\top})
\]
and then we solve for rotation matrix $R$. Plugging it back into the $t^*$ equations we get $t$.

To train the WeightedPose, we use a set of losses defined below, which are similar to those in \cite{pan2022tax}. We assume we have access to a set of demonstrations of the task, in which the action and anchor objects are in the target relative pose such that $\TAB = \mathbf{I}$.

\textbf{Point Displacement Loss:}

Rather than directly supervising the rotation and translation (as seen in DCP), we manage the predicted transformation by observing its impact on the points. To do this, we utilize the point clouds from the objects in their demonstration setup and apply a random transformation to each cloud for this loss, $\mathbf{\hat{P}}_{\mathcal{A}} = \mathbf{T}_{\alpha} \mathbf{P}_{\mathcal{A}}$, and $\mathbf{\hat{P}}_{\mathcal{B}} = \mathbf{T}_{\beta} \mathbf{P}_{\mathcal{B}}$. This would give us a ground truth transform of $\TAB^{GT} =\mathbf{T}_{\beta}\mathbf{T}_{\alpha}^{-1}$;
the inverse of this transform would move object $\mathcal{B}$ to the correct position relative to object $\mathcal{A}$. Using the actual transformation as a reference, we calculate the Mean Squared Error (MSE) loss by comparing the correctly transformed points to those transformed based on our predicted values. 
\begin{equation}
    \mathcal{L}_\mathrm{disp} = \left\|\mathbf{T}_{\mathcal{A}\mathcal{B}} \mathbf{P}_{\mathcal{A}}  - \mathbf{T}_{\mathcal{A}\mathcal{B}}^{GT} \mathbf{P}_{\mathcal{A}}\right\|^2 + \left\|\mathbf{T}_{\mathcal{A}\mathcal{B}}^{-1} \mathbf{P}_{\mathcal{B}}  - \mathbf{T}_{\mathcal{A}\mathcal{B}}^{ GT-1} \mathbf{P}_{\mathcal{B}}\right\|^2
    \label{equ:disp_loss}
\end{equation}
\textbf{Direct Correspondence Loss.}
Although the Point Displacement Loss accurately reflects errors observed during inference, it may result in correspondences that are individually inaccurate yet average out to the correct pose. To enhance these errors, we directly supervise the correspondences that are learned $\tilde{V}_{\mathcal{A}}$ and $\tilde{V}_{\mathcal{B}}$: 
\begin{equation}
    \mathcal{L}_\mathrm{corr} = \left\| \VA - \mathbf{T}_{\mathcal{A}\mathcal{B}}^{GT} \mathbf{P}_{\mathcal{A}}\right\|^2 + \left\|\VB - \mathbf{T}_{\mathcal{A}\mathcal{B}}^{GT-1} \mathbf{P}_{\mathcal{B}}\right\|^2.
    \label{equ:corr_loss}
\end{equation}
\textbf{Correspondence Consistency Loss.}
Additionally, a consistency loss may be employed. This loss penalizes any discrepancies between correspondences and the final predicted transformation. An advantage of this loss is its ability to encourage the network to maintain object rigidity while it learns precise object placement. It is important to note that this is akin to the Direct Correspondence Loss, except that it relies on the predicted transformation rather than the ground truth. Therefore, this loss does not require any ground truth data:
\begin{equation}
    \mathcal{L}_\mathrm{cons} =  \left\| \VA - \mathbf{T}_{\mathcal{A}\mathcal{B}} \mathbf{P}_{\mathcal{A}}\right\|^2 + \left\|\VB - \mathbf{T}_{\mathcal{A}\mathcal{B}}^{-1} \mathbf{P}_{\mathcal{B}}\right\|^2.
    \label{equ:cons_loss}
\end{equation}
\textbf{Direct SE(3) Transformation Loss. }We also define a loss directly for the transformation outputted in $SE(3)$, supervising between the SVD output pose and ground-truth pose. 
\begin{equation}
    \mathcal{L}_\text{tf} = ||\mathbf{T}_{\mathcal{A}\mathcal{B}} - \mathbf{T}_{\mathcal{A}\mathcal{B}}^{GT}||_F
\end{equation}
\section{Experiments}

We describe a PartNet-Mobility Placement task as the act of positioning an action object in relation to an anchor object according to a predefined semantic goal position. This task demands that the model generate a cross-pose for both articulated parts and free-floating objects. We have chosen a collection of household furniture items from the PartNet-Mobility dataset for this purpose~\cite{Xiang2020-oz} and a set of small rigid objects released with the Ravens simulation environment \cite{zeng2020transporter}. When the model is required to estimate goal pose for the articulated part, following the terminology in \cite{pan2022tax}, articulated part serves as the action object, while the static body of the furniture acts as the anchor object. Conversely, when the model estimates the goal pose for free-floating objects, the scenario aligns with the same setting as previously discussed \cite{pan2022tax}. 
\begin{figure}
    \centering
    \includegraphics[width=0.7\textwidth]{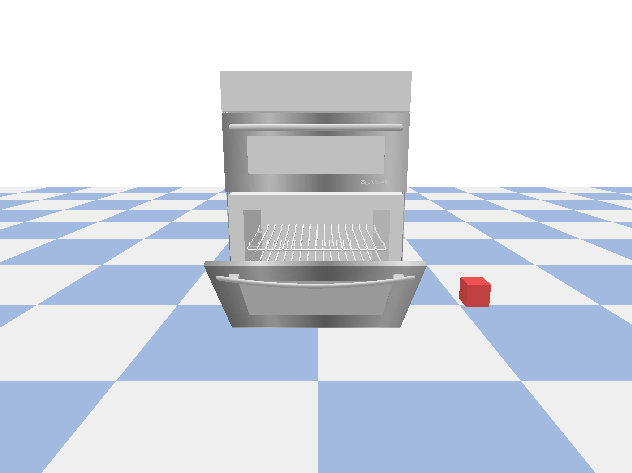}
    \caption{Task illustration. The model needs to first output goal pose for opening the oven door, and then output goal pose for putting the block inside the oven.}
    \label{fig:my_label}
\end{figure}
We compare the proposed method to several baselines. First, we compare against TAX-Pose trained on both free-floating and articulated objects as well as Goal-Flow trained on both categories of objects. We then compare against an oracle model that classifies if the objects of interest are free-floating or articulated, and deploys the model accordingly. 

We describe the baselines in details:
\begin{itemize}
    \item Goal Flow Pretrained (GF Pretrained): Pretrain Goal Flow on articulated objects only and test on both articulated and free-floating objects.
    \item TAX-Pose Pretrained (TP Pretrained): Pretrain TAX-Pose on free-floating objects only and test on both articulated and free-floating objects.
    \item Weighted Pose Original Loss (WP OG Loss): Weighted Pose trained and  test  on both free-floating and articulated objects using the original TAX-Pose loss.
    \item  Weighted Pose Post-SVD Loss (WP Post-SVD): Weighted Pose trained and  test  on both free-floating and articulated objects but using the post-SVD TAX-Pose loss.
    \item Weighted Pose Post-SVD \& Transformation Loss (WP Post-SVD): Weighted Pose trained and  test  on both free-floating and articulated objects but using both the post-SVD TAX-Pose loss and direct transformation loss. Where the transformation loss is the MSE between the predicted $SE(3)$ transformation and the ground-truth $SE(3)$ transformation.
\end{itemize}

We use the same PartNet-Mobility dataset as in TAX-Pose to evaluate the system. For semantic tasks, we will only select the ``In'' task as it is the most intuitive task. However, other than just performing evaluation on the free-floating objects, we will also evaluate the method on opening the articulate objects. Under this formulation, the action object is the articulated part and the anchor object is the static body. We will measure rotational error. translational error, and per-point MSE in our experiments. We follow the same train-val split as in TAX-Pose's PartNet-Mobility dataset. 

In Table \ref{tab:baselines-sr}, Goal Flow Pretrained baseline fails on free-floating objects and excels on articulated objects. Similarly for TAX-Pose pretrained, since it was trained on free-floating objects, it does do well on articulated objects, as the high rotational and translational errors indicate.

We next evaluate the variations of WeightedPose, which differ in the training paradigms. Using Weighted Pose Original Loss, which is trained using the original TAX-Pose loss, we are able to achieve better performance on both articulated and free-floating objects. Interestingly, the results we achieve using Weighted Pose Original Loss for articulated objects are better than the results from using Goal Flow pretrained on articulated objects.

While ideally, the $w$ learned in this method should effectively act as a classifier of the input object (1 for articulated objects, 0 for free-floating objects). Intuitively, given a perfect $w$, the performance of WeightedPose would be upper-bounded by Goal Flow's performance on articulated objects and by TAX-Pose on free-floating objects. However, a weighted combination of the two results due to the imperfect learned $w$ weight could potentially correct the mistake made by each model by summing the results with a weighted sum. This may explain why WeightedPose sometimes performs better than its hypothesized upper bound on some objects.

Lastly, results suggest that the original loss used in TAX-Pose yields the best overall performance. However, it is worth noting that by using post-SVD loss during training, we are able to achieve lower translational error in test time. Interestingly, by introducing a direct $SE(3)$ supervision, the results degrade marginally.


\begin{table}[ht]
\resizebox{\textwidth}{!}{
\begin{tabular}{|l|l|rr|rr|rr|rr|rr|}
\hline
 &  & \multicolumn{2}{c|}{\textbf{GF Pretrained}} & \multicolumn{2}{c|}{\textbf{TP Pretrained}} & \multicolumn{2}{c|}{\textbf{WP OG Loss}} & \multicolumn{2}{c|}{\textbf{WP Post-SVD}} & \multicolumn{2}{c|}{\textbf{WP Post SVD + T}} \\ \cline{2-12} 
 & \textbf{Metrics} & \multicolumn{1}{l|}{FF} & \multicolumn{1}{l|}{Art} & \multicolumn{1}{l|}{FF} & \multicolumn{1}{l|}{Art} & \multicolumn{1}{l|}{FF} & \multicolumn{1}{l|}{Art} & \multicolumn{1}{l|}{FF} & \multicolumn{1}{l|}{Art} & \multicolumn{1}{l|}{FF} & \multicolumn{1}{l|}{Art} \\ \hline
\multirow{3}{*}{\textbf{Train}} & \textbf{Rot err} & \multicolumn{1}{r|}{31.61} & \textbf{5.16} & \multicolumn{1}{r|}{\textbf{2.61}} & 55.78 & \multicolumn{1}{r|}{11.69} & 3.15 & \multicolumn{1}{r|}{13.14} & 3.28 & \multicolumn{1}{r|}{14.04} & 5.52 \\
 & \textbf{Trans err} & \multicolumn{1}{r|}{1.21} & \textbf{0.16} & \multicolumn{1}{r|}{\textbf{0.04}} & 0.98 & \multicolumn{1}{r|}{0.14} & 0.08 & \multicolumn{1}{r|}{0.21} & 0.07 & \multicolumn{1}{r|}{0.21} & 0.07 \\
 & \textbf{PP MSE} & \multicolumn{1}{r|}{1.04} & \textbf{0.04} & \multicolumn{1}{r|}{\textbf{0.01}} & 0.85 & \multicolumn{1}{r|}{0.07} & 0.05 & \multicolumn{1}{r|}{0.09} & 0.04 & \multicolumn{1}{r|}{0.11} & 0.08 \\ \hline
\multirow{3}{*}{\textbf{Val}} & \textbf{Rot err} & \multicolumn{1}{r|}{35.5} & \textbf{9.14} & \multicolumn{1}{r|}{\textbf{9.87}} & 59.73 & \multicolumn{1}{r|}{11.22} & 9.01 & \multicolumn{1}{r|}{14.13} & 9.71 & \multicolumn{1}{r|}{13.06} & 10.03 \\
 & \textbf{Trans err} & \multicolumn{1}{r|}{1.3} & \textbf{0.19} & \multicolumn{1}{r|}{\textbf{0.18}} & 0.99 & \multicolumn{1}{r|}{0.26} & 0.15 & \multicolumn{1}{r|}{0.22} & 0.18 & \multicolumn{1}{r|}{0.23} & 0.18 \\
 & \textbf{PP MSE} & \multicolumn{1}{r|}{1.07} & \textbf{0.1} & \multicolumn{1}{r|}{\textbf{0.15}} & 0.82 & \multicolumn{1}{r|}{0.16} & 0.11 & \multicolumn{1}{r|}{0.17} & 0.12 & \multicolumn{1}{r|}{0.18} & 0.11 \\ \hline
\end{tabular}
}
\caption{Weighted Pose Results: We compare Rotation Error, Translational Error, and Per-Point MSE for both training and validation objects. }
\vspace*{-10pt}
\label{tab:baselines-sr}
\end{table}

\section{Conclusions and Future Work}

In conclusion, we have presented a method to combine the two architectures using weighted SVD. While this model is more of a proof-of-concept that attempts to unify the two architectures, it is worth pointing out that using the two models for the two categories of objects is able to help us generate goal poses for various free-floating and articulated objects. Moreover, by finetuning pretrained models from a weighted SVD combination, we are able to outperform the models on their respective training datasets categories. In future work, we wish to generalize the mathematics of the combined architecture beyond tasks that involve articulated and free-floating objects. We would also like to explore how such a unified architecture can aid motion planning as a geometric suggestor.

\bibliographystyle{plainnat}
\bibliography{ref}

\begin{thebibliography}{45}
\providecommand{\natexlab}[1]{#1}
\providecommand{\url}[1]{\texttt{#1}}
\expandafter\ifx\csname urlstyle\endcsname\relax
  \providecommand{\doi}[1]{doi: #1}\else
  \providecommand{\doi}{doi: \begingroup \urlstyle{rm}\Url}\fi

\bibitem[Avigal et~al.(2020)Avigal, Paradis, and Zhang]{avigal20206}
Yahav Avigal, Samuel Paradis, and Harry Zhang.
\newblock 6-dof grasp planning using fast 3d reconstruction and grasp quality
  cnn.
\newblock \emph{arXiv preprint arXiv:2009.08618}, 2020.

\bibitem[Avigal et~al.(2021)Avigal, Satish, Tam, Huang, Zhang, Danielczuk,
  Ichnowski, and Goldberg]{avigal2021avplug}
Yahav Avigal, Vishal Satish, Zachary Tam, Huang Huang, Harry Zhang, Michael
  Danielczuk, Jeffrey Ichnowski, and Ken Goldberg.
\newblock Avplug: Approach vector planning for unicontact grasping amid
  clutter.
\newblock In \emph{2021 IEEE 17th International Conference on Automation
  Science and Engineering (CASE)}, pages 1140--1147. IEEE, 2021.

\bibitem[Berenson et~al.(2011)Berenson, Srinivasa, and
  Kuffner]{berenson2011task}
Dmitry Berenson, Siddhartha Srinivasa, and James Kuffner.
\newblock Task space regions: A framework for pose-constrained manipulation
  planning.
\newblock \emph{The International Journal of Robotics Research}, 30\penalty0
  (12):\penalty0 1435--1460, 2011.

\bibitem[Burget et~al.(2013)Burget, Hornung, and Bennewitz]{Burget2013-nb}
Felix Burget, Armin Hornung, and Maren Bennewitz.
\newblock Whole-body motion planning for manipulation of articulated objects.
\newblock In \emph{2013 {IEEE} International Conference on Robotics and
  Automation}, pages 1656--1662, May 2013.

\bibitem[Cheong et~al.(2007)Cheong, Van Der~Stappen, Goldberg, Overmars, and
  Rimon]{Cheong2007-iw}
Jae-Sook Cheong, A~Frank Van Der~Stappen, Ken Goldberg, Mark~H Overmars, and
  Elon Rimon.
\newblock {Immobilizing} {Hinged} {Polygons}.
\newblock \emph{Int. J. Comput. Geom. Appl.}, 17\penalty0 (01):\penalty0
  45--69, February 2007.

\bibitem[Chitta et~al.(2010)Chitta, Cohen, and Likhachev]{Chitta2010-vn}
Sachin Chitta, Benjamin Cohen, and Maxim Likhachev.
\newblock Planning for autonomous door opening with a mobile manipulator.
\newblock In \emph{2010 {IEEE} International Conference on Robotics and
  Automation}, pages 1799--1806, May 2010.

\bibitem[Devgon et~al.(2020)Devgon, Ichnowski, Balakrishna, Zhang, and
  Goldberg]{devgon2020orienting}
Shivin Devgon, Jeffrey Ichnowski, Ashwin Balakrishna, Harry Zhang, and Ken
  Goldberg.
\newblock Orienting novel 3d objects using self-supervised learning of rotation
  transforms.
\newblock In \emph{2020 IEEE 16th International Conference on Automation
  Science and Engineering (CASE)}, pages 1453--1460. IEEE, 2020.

\bibitem[Elmquist et~al.(2022)Elmquist, Young, Hansen, Ashokkumar, Caldararu,
  Dashora, Mahajan, Zhang, Fang, Shen, et~al.]{elmquist2022art}
Asher Elmquist, Aaron Young, Thomas Hansen, Sriram Ashokkumar, Stefan
  Caldararu, Abhiraj Dashora, Ishaan Mahajan, Harry Zhang, Luning Fang,
  He~Shen, et~al.
\newblock Art/atk: A research platform for assessing and mitigating the
  sim-to-real gap in robotics and autonomous vehicle engineering.
\newblock \emph{arXiv preprint arXiv:2211.04886}, 2022.

\bibitem[Florence et~al.(2018)Florence, Manuelli, and
  Tedrake]{florence2018dense}
Peter~R Florence, Lucas Manuelli, and Russ Tedrake.
\newblock Dense object nets: Learning dense visual object descriptors by and
  for robotic manipulation.
\newblock In \emph{Conference on Robot Learning}, pages 373--385. PMLR, 2018.

\bibitem[He et~al.(2020)He, Sun, Huang, Liu, Fan, and Sun]{he2020pvn3d}
Yisheng He, Wei Sun, Haibin Huang, Jianran Liu, Haoqiang Fan, and Jian Sun.
\newblock Pvn3d: A deep point-wise 3d keypoints voting network for 6dof pose
  estimation.
\newblock In \emph{Proceedings of the IEEE/CVF conference on computer vision
  and pattern recognition}, pages 11632--11641, 2020.

\bibitem[He et~al.(2021)He, Huang, Fan, Chen, and Sun]{he2021ffb6d}
Yisheng He, Haibin Huang, Haoqiang Fan, Qifeng Chen, and Jian Sun.
\newblock Ffb6d: A full flow bidirectional fusion network for 6d pose
  estimation.
\newblock In \emph{Proceedings of the IEEE/CVF Conference on Computer Vision
  and Pattern Recognition}, pages 3003--3013, 2021.

\bibitem[Hu et~al.(2017)Hu, Li, Van~Kaick, Shamir, Zhang, and Huang]{Hu2017-bn}
Ruizhen Hu, Wenchao Li, Oliver Van~Kaick, Ariel Shamir, Hao Zhang, and Hui
  Huang.
\newblock Learning to predict part mobility from a single static snapshot.
\newblock \emph{ACM Trans. Graph.}, 36\penalty0 (6):\penalty0 1--13, November
  2017.

\bibitem[Jain et~al.(2021)Jain, Lioutikov, Chuck, and Niekum]{Jain2021-rg}
Ajinkya Jain, Rudolf Lioutikov, Caleb Chuck, and Scott Niekum.
\newblock {ScrewNet}: {Category-Independent} articulation model estimation from
  depth images using screw theory.
\newblock In \emph{2021 {IEEE} International Conference on Robotics and
  Automation ({ICRA})}, pages 13670--13677, May 2021.

\bibitem[Jin et~al.(2024)Jin, Karmalkar, Zhang, and Carlone]{jin2024multi}
David Jin, Sushrut Karmalkar, Harry Zhang, and Luca Carlone.
\newblock Multi-model 3d registration: Finding multiple moving objects in
  cluttered point clouds.
\newblock \emph{arXiv preprint arXiv:2402.10865}, 2024.

\bibitem[Katz et~al.(2008)Katz, Pyuro, and Brock]{Katz2008-jo}
Dov Katz, Yuri Pyuro, and Oliver Brock.
\newblock Learning to manipulate articulated objects in unstructured
  environments using a grounded relational representation.
\newblock In \emph{Robotics: Science and Systems {IV}}. Robotics: Science and
  Systems Foundation, June 2008.

\bibitem[Li et~al.(2020)Li, Wang, Yi, Guibas, Lynn~Abbott, and Song]{Li2020-go}
Xiaolong Li, He~Wang, Li~Yi, Leonidas~J Guibas, A~Lynn~Abbott, and Shuran Song.
\newblock {Category-Level} articulated object pose estimation, 2020.

\bibitem[Lim et~al.(2021)Lim, Huang, Chen, Wang, Ichnowski, Seita, Laskey, and
  Goldberg]{lim2021planar}
Vincent Lim, Huang Huang, Lawrence~Yunliang Chen, Jonathan Wang, Jeffrey
  Ichnowski, Daniel Seita, Michael Laskey, and Ken Goldberg.
\newblock Planar robot casting with real2sim2real self-supervised learning.
\newblock \emph{arXiv preprint arXiv:2111.04814}, 2021.

\bibitem[Lim et~al.(2022)Lim, Huang, Chen, Wang, Ichnowski, Seita, Laskey, and
  Goldberg]{lim2022real2sim2real}
Vincent Lim, Huang Huang, Lawrence~Yunliang Chen, Jonathan Wang, Jeffrey
  Ichnowski, Daniel Seita, Michael Laskey, and Ken Goldberg.
\newblock Real2sim2real: Self-supervised learning of physical single-step
  dynamic actions for planar robot casting.
\newblock In \emph{2022 International Conference on Robotics and Automation
  (ICRA)}, pages 8282--8289. IEEE, 2022.

\bibitem[Lowe(1999)]{lowe1999object}
David~G Lowe.
\newblock Object recognition from local scale-invariant features.
\newblock In \emph{Proceedings of the seventh IEEE international conference on
  computer vision}, volume~2, pages 1150--1157. Ieee, 1999.

\bibitem[Manuelli et~al.(2019)Manuelli, Gao, Florence, and
  Tedrake]{manuelli2019kpam}
Lucas Manuelli, Wei Gao, Peter Florence, and Russ Tedrake.
\newblock kpam: Keypoint affordances for category-level robotic manipulation.
\newblock \emph{International Symposium on Robotics Research (ISRR) 2019},
  2019.

\bibitem[Manuelli et~al.(2021)Manuelli, Li, Florence, and
  Tedrake]{manuelli2021keypoints}
Lucas Manuelli, Yunzhu Li, Pete Florence, and Russ Tedrake.
\newblock Keypoints into the future: Self-supervised correspondence in
  model-based reinforcement learning.
\newblock In \emph{Conference on Robot Learning}, pages 693--710. PMLR, 2021.

\bibitem[Mo et~al.(2019)Mo, Zhu, Chang, Yi, Tripathi, Guibas, and
  Su]{Mo2019-az}
Kaichun Mo, Shilin Zhu, Angel~X Chang, Li~Yi, Subarna Tripathi, Leonidas~J
  Guibas, and Hao Su.
\newblock Partnet: A large-scale benchmark for fine-grained and hierarchical
  part-level 3d object understanding.
\newblock In \emph{Proceedings of the {IEEE/CVF} Conference on Computer Vision
  and Pattern Recognition}, pages 909--918, 2019.

\bibitem[Mo et~al.(2021)Mo, Guibas, Mukadam, Gupta, and Tulsiani]{Mo2021-jm}
Kaichun Mo, Leonidas~J Guibas, Mustafa Mukadam, Abhinav Gupta, and Shubham
  Tulsiani.
\newblock Where2act: From pixels to actions for articulated 3d objects.
\newblock In \emph{Proceedings of the IEEE/CVF International Conference on
  Computer Vision}, pages 6813--6823, 2021.

\bibitem[Narayanan and Likhachev(2015)]{Narayanan2015-mp}
Venkatraman Narayanan and Maxim Likhachev.
\newblock Task-oriented planning for manipulating articulated mechanisms under
  model uncertainty.
\newblock In \emph{2015 {IEEE} International Conference on Robotics and
  Automation ({ICRA})}, pages 3095--3101, May 2015.

\bibitem[Pan et~al.(2022)Pan, Okorn, Zhang, Eisner, and Held]{pan2022tax}
Chuer Pan, Brian Okorn, Harry Zhang, Ben Eisner, and David Held.
\newblock Tax-pose: Task-specific cross-pose estimation for robot manipulation.
\newblock \emph{arXiv preprint arXiv:2211.09325}, 2022.

\bibitem[Pan et~al.(2023)Pan, Okorn, Zhang, Eisner, and Held]{pan2023tax}
Chuer Pan, Brian Okorn, Harry Zhang, Ben Eisner, and David Held.
\newblock Tax-pose: Task-specific cross-pose estimation for robot manipulation.
\newblock In \emph{Conference on Robot Learning}, pages 1783--1792. PMLR, 2023.

\bibitem[Qin et~al.(2020)Qin, Fang, Zhu, Fei-Fei, and Savarese]{qin2020keto}
Zengyi Qin, Kuan Fang, Yuke Zhu, Li~Fei-Fei, and Silvio Savarese.
\newblock Keto: Learning keypoint representations for tool manipulation.
\newblock In \emph{2020 IEEE International Conference on Robotics and
  Automation (ICRA)}, pages 7278--7285. IEEE, 2020.

\bibitem[Rothganger et~al.(2006)Rothganger, Lazebnik, Schmid, and
  Ponce]{rothganger20063d}
Fred Rothganger, Svetlana Lazebnik, Cordelia Schmid, and Jean Ponce.
\newblock 3d object modeling and recognition using local affine-invariant image
  descriptors and multi-view spatial constraints.
\newblock \emph{International journal of computer vision}, 66\penalty0
  (3):\penalty0 231--259, 2006.

\bibitem[Shen et~al.(2024)Shen, Zhu, Fan, Zhang, and Wu]{shen2024diffclip}
Sitian Shen, Zilin Zhu, Linqian Fan, Harry Zhang, and Xinxiao Wu.
\newblock Diffclip: Leveraging stable diffusion for language grounded 3d
  classification.
\newblock In \emph{Proceedings of the IEEE/CVF Winter Conference on
  Applications of Computer Vision}, pages 3596--3605, 2024.

\bibitem[Sim et~al.(2019)Sim, Beaufays, Benard, Guliani, Kabel, Khare,
  Lucassen, Zadrazil, Zhang, Johnson, et~al.]{sim2019personalization}
Khe~Chai Sim, Fran{\c{c}}oise Beaufays, Arnaud Benard, Dhruv Guliani, Andreas
  Kabel, Nikhil Khare, Tamar Lucassen, Petr Zadrazil, Harry Zhang, Leif
  Johnson, et~al.
\newblock Personalization of end-to-end speech recognition on mobile devices
  for named entities.
\newblock In \emph{2019 IEEE Automatic Speech Recognition and Understanding
  Workshop (ASRU)}, pages 23--30. IEEE, 2019.

\bibitem[Simeonov et~al.(2022)Simeonov, Du, Tagliasacchi, Tenenbaum, Rodriguez,
  Agrawal, and Sitzmann]{simeonov2021neural}
Anthony Simeonov, Yilun Du, Andrea Tagliasacchi, Joshua~B Tenenbaum, Alberto
  Rodriguez, Pulkit Agrawal, and Vincent Sitzmann.
\newblock Neural descriptor fields: Se (3)-equivariant object representations
  for manipulation.
\newblock In \emph{2022 International Conference on Robotics and Automation
  (ICRA)}, pages 6394--6400. IEEE, 2022.

\bibitem[Turpin et~al.(2021)Turpin, Wang, Tsogkas, Dickinson, and
  Garg]{turpin2021gift}
Dylan Turpin, Liquan Wang, Stavros Tsogkas, Sven Dickinson, and Animesh Garg.
\newblock Gift: Generalizable interaction-aware functional tool affordances
  without labels.
\newblock \emph{Robotics: Science and Systems (RSS)}, 2021.

\bibitem[Vecerik et~al.(2021)Vecerik, Regli, Sushkov, Barker, Pevceviciute,
  Roth{\"o}rl, Hadsell, Agapito, and Scholz]{vecerik2021s3k}
Mel Vecerik, Jean-Baptiste Regli, Oleg Sushkov, David Barker, Rugile
  Pevceviciute, Thomas Roth{\"o}rl, Raia Hadsell, Lourdes Agapito, and Jonathan
  Scholz.
\newblock S3k: Self-supervised semantic keypoints for robotic manipulation via
  multi-view consistency.
\newblock In \emph{Conference on Robot Learning}, pages 449--460. PMLR, 2021.

\bibitem[Wang et~al.(2019)Wang, Zhou, Shi, Chen, Zhao, and Xu]{Wang2019-gy}
Xiaogang Wang, Bin Zhou, Yahao Shi, Xiaowu Chen, Qinping Zhao, and Kai Xu.
\newblock Shape2motion: Joint analysis of motion parts and attributes from 3d
  shapes.
\newblock In \emph{Proceedings of the {IEEE/CVF} Conference on Computer Vision
  and Pattern Recognition}, pages 8876--8884, 2019.

\bibitem[Xiang et~al.(2020)Xiang, Qin, Mo, Xia, Zhu, Liu, Liu, Jiang, Yuan,
  Wang, and {Others}]{Xiang2020-oz}
Fanbo Xiang, Yuzhe Qin, Kaichun Mo, Yikuan Xia, Hao Zhu, Fangchen Liu, Minghua
  Liu, Hanxiao Jiang, Yifu Yuan, He~Wang, and {Others}.
\newblock Sapien: A simulated part-based interactive environment.
\newblock In \emph{Proceedings of the {IEEE/CVF} Conference on Computer Vision
  and Pattern Recognition}, pages 11097--11107, 2020.

\bibitem[Xiang et~al.(2018)Xiang, Schmidt, Narayanan, and
  Fox]{xiang2017posecnn}
Yu~Xiang, Tanner Schmidt, Venkatraman Narayanan, and Dieter Fox.
\newblock Posecnn: A convolutional neural network for 6d object pose estimation
  in cluttered scenes.
\newblock \emph{Robotics: Science and Systems (RSS)}, 2018.

\bibitem[Xu et~al.(2022)Xu, Zhanpeng, and Song]{Xu2021-iw}
Zhenjia Xu, He~Zhanpeng, and Shuran Song.
\newblock Umpnet: Universal manipulation policy network for articulated
  objects.
\newblock \emph{IEEE Robotics and Automation Letters}, 2022.

\bibitem[Yan et~al.(2020)Yan, Hu, Yan, Chen, Van~Kaick, Zhang, and
  Huang]{Yan2020-hm}
Zihao Yan, Ruizhen Hu, Xingguang Yan, Luanmin Chen, Oliver Van~Kaick, Hao
  Zhang, and Hui Huang.
\newblock Rpm-net: recurrent prediction of motion and parts from point cloud.
\newblock \emph{arXiv preprint arXiv:2006.14865}, 2020.

\bibitem[Yao et~al.(2023)Yao, Deng, Cao, Zhang, and Deng]{yao2023apla}
Yupu Yao, Shangqi Deng, Zihan Cao, Harry Zhang, and Liang-Jian Deng.
\newblock Apla: Additional perturbation for latent noise with adversarial
  training enables consistency.
\newblock \emph{arXiv preprint arXiv:2308.12605}, 2023.

\bibitem[Zeng et~al.(2021)Zeng, Florence, Tompson, Welker, Chien, Attarian,
  Armstrong, Krasin, Duong, Sindhwani, et~al.]{zeng2020transporter}
Andy Zeng, Pete Florence, Jonathan Tompson, Stefan Welker, Jonathan Chien,
  Maria Attarian, Travis Armstrong, Ivan Krasin, Dan Duong, Vikas Sindhwani,
  et~al.
\newblock Transporter networks: Rearranging the visual world for robotic
  manipulation.
\newblock In \emph{Conference on Robot Learning}, pages 726--747. PMLR, 2021.

\bibitem[Zeng et~al.(2020)Zeng, Lee, Liang, and Kroemer]{Zeng2020-tk}
Vicky Zeng, Timothy~E Lee, Jacky Liang, and Oliver Kroemer.
\newblock Visual identification of articulated object parts.
\newblock In \emph{2021 IEEE/RSJ International Conference on Intelligent Robots
  and Systems (IROS)}, pages 2443--2450. IEEE, 2020.

\bibitem[Zhang(2016)]{zhang2016health}
Haolun Zhang.
\newblock Health diagnosis based on analysis of data captured by wearable
  technology devices.
\newblock \emph{International Journal of Advanced Science and Technology},
  95:\penalty0 89--96, 2016.

\bibitem[Zhang et~al.(2020)Zhang, Ichnowski, Avigal, Gonzales, Stoica, and
  Goldberg]{zhang2020dex}
Harry Zhang, Jeffrey Ichnowski, Yahav Avigal, Joseph Gonzales, Ion Stoica, and
  Ken Goldberg.
\newblock Dex-net ar: Distributed deep grasp planning using a commodity
  cellphone and augmented reality app.
\newblock In \emph{2020 IEEE International Conference on Robotics and
  Automation (ICRA)}, pages 552--558. IEEE, 2020.

\bibitem[Zhang et~al.(2021)Zhang, Ichnowski, Seita, Wang, Huang, and
  Goldberg]{zhang2021robots}
Harry Zhang, Jeffrey Ichnowski, Daniel Seita, Jonathan Wang, Huang Huang, and
  Ken Goldberg.
\newblock Robots of the lost arc: Self-supervised learning to dynamically
  manipulate fixed-endpoint cables.
\newblock In \emph{2021 IEEE International Conference on Robotics and
  Automation (ICRA)}, pages 4560--4567. IEEE, 2021.

\bibitem[Zhang et~al.(2023)Zhang, Eisner, and Held]{zhang2023flowbot++}
Harry Zhang, Ben Eisner, and David Held.
\newblock Flowbot++: Learning generalized articulated objects manipulation via
  articulation projection.
\newblock \emph{arXiv preprint arXiv:2306.12893}, 2023.

\end{thebibliography}
\end{document}